\documentclass[journal]{IEEEtran}

\usepackage[cmex10]{amsmath}
\usepackage{amsthm}
\usepackage{amssymb}
\usepackage{mathrsfs}
\usepackage{graphicx}
\usepackage{float}
\usepackage{array}
\usepackage{epstopdf}
\usepackage{multirow}
\usepackage[justification=centering]{caption}

\onecolumn

\begin{document}

\title{A Machine Learning Approach For Identifying Patients with Mild Traumatic Brain Injury Using Diffusion MRI Modeling}

\author{Shervin~Minaee$^{1,*}$, Yao~Wang$^1$, Sohae Chung$^2$, Xiuyuan Wang$^2$, Els Fieremans$^2$, Steven Flanagan$^3$, Joseph Rath$^3$, Yvonne W. Lui$^2$,  \\
$^1$Electrical and Computer Engineering Department, New York University
\\ $^2$Department of Radiology, New York University
\\ $^3$Department of Rehabilitation Medicine, New York University
\\ $^*$Correspondence to Shervin Minaee: shervin.minaee@nyu.edu}

\maketitle

\IEEEpeerreviewmaketitle

\vspace{-0.5cm}

\noindent
\textbf{Purpose.}
While diffusion MRI has been extremely promising in the study of MTBI, identifying patients with recent MTBI remains a challenge \cite{yvonne}. The literature is mixed with regard to localizing injury in these patients, however, gray matter such as the thalamus and white matter including the corpus callosum and frontal deep white matter have been repeatedly implicated as areas at high risk for injury. The purpose of this study is to develop a machine learning framework to classify MTBI patients and controls using features derived from multi-shell diffusion MRI in the thalamus, frontal white matter and corpus callosum.
\\

\noindent
\textbf{Method.}
This IRB-approved, prospective study includes 69 MTBI subjects between 18 and 64 years old, within 1 month of MTBI as defined by the American College of Rehabilitation Medicine (ACRM) criteria for head injury and 40 healthy age and sex-matched controls. Imaging was performed on a 3.0 Tesla Siemens Trio (Erlangen, Germany) magnet including multi-shell diffusion MRI at 5 b-values (250, 1000, 1500, 2000, 2500 s/mm2) in a total of 136 directions using multiband 2 at isotropic 2.5mm image resolution.
Denoising, Gibbs correction, geometric and eddy current distortion and motion correction was performed. In addition to Mean Diffusion (MD), Fractional Anisotropy (FA) and Mean Kurtosis (MK), additional modeled white matter tract integrity (WMTI) metrics were calculated and parametric maps generated for: 1) axonal water fraction (AWF), 2) intra-axonal diffusivity (``D''\_``axon'', diffusivity within axons), 3) extra-axonal axial and 4) extra-axonal radial diffusivities (``De''\_``par''  and ``De''\_``per'', diffusion parallel and perpendicular to the axonal tracts in the extra-axonal space). The relationship between WMTI metrics and tissue microstructure has been studied in several animal validation works, specifically demyelination following acute injury and remyelination in the chronic phase and also applied in vivo to characterize normal and abnormal neural development in the central nervous system (CNS).
\\Two approaches were attempted differing in feature representation. In the first approach, mean values of above 7 MR metrics in five regions (thalamus, prefrontal white-matter, corpus callosum (CC) Body, CC-Genu, and CC-Splenium) were used in the feature representation along with two demographic features (age and sex) and four neurocognitive tests (Stroop, Symbol Digit Modalities Test (SDMT), California Verbal Learning Test (CVLT) and Fatigue Severity Scale (FSS)). A  greedy forward feature selection approach was used to choose the best feature subset upon which a support vector machine achieves the highest cross-validation accuracy. 
\\In the second approach, 16x16 patches are extracted from brain slices through the areas of interest and all the training patches in the MTBI patients and control subjects are separately clustered to learn the most representative visual patterns (called “visual” words). Each 3D dataset is then shown as a histogram of these words, known as the Bag of Words (BoW) representation \cite{bow}. The block-diagram of this approach is shown in Fig. 1. The same feature selection approach was used to derive a best feature subset for the support vector machine classifier \cite{svm}.

\begin{figure}[h]
\begin{center}
    \includegraphics [scale=0.395] {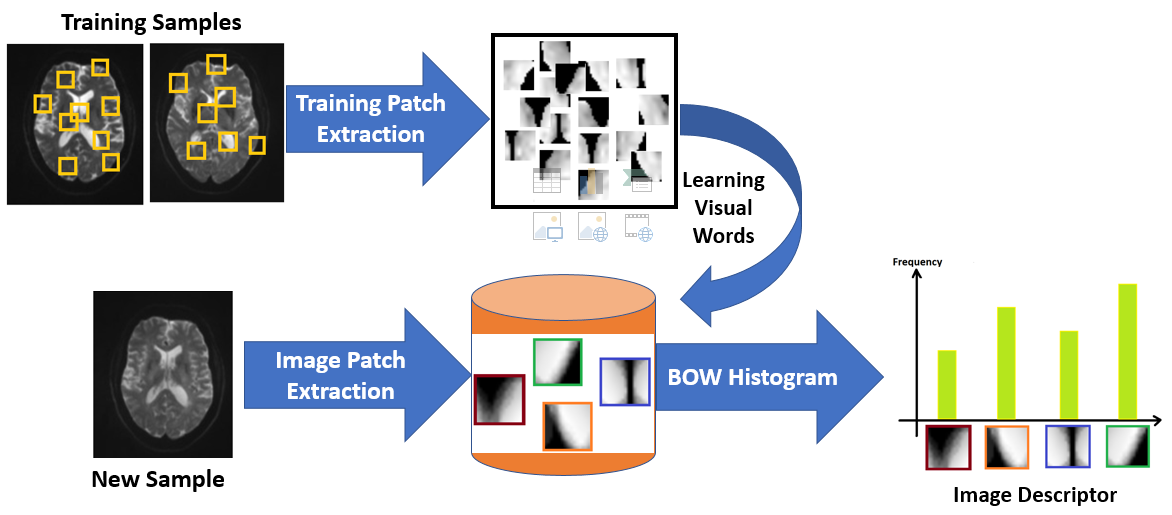}
\end{center}
\vspace{-0.2cm}
  \caption{Schematic of ``Bag of Words'' feature extraction approach.}
\end{figure}

\noindent
\textbf{Results.}
Using approach 1, we found the best single metric cross-validation classification accuracy of 72\% using AWF in CC-Body. The best feature subset chosen by the greedy feature selection had an accuracy of 80\% with 8 features (De\_par in thalamus, De\_par and DA in pre-frontal white matter, FA in CC-Genu, AWF and De\_per in CC-Body, Stroop and SDMT).
\\BoW approach achieved further improvement in accuracy to 85.5\%. In this approach, the raw representation is 206 dimensional, including 20 words for each of 5 MR metrics (AWF, DA, De\_par, FA, MD) in each of 2 brain regions (thalamus and corpus collasum), and 6 demographic and neurocognitive features, and the optimum subset contains 8 features (which includes age, and words from AWF, De\_par  and FA in thalamus, and also DA, De\_par, FA and MD  in corpus callosum). Fig. 2 shows example BoW histograms. 
We can see MTBI and control subjects have clear differences in frequency of some visual words, for example along the right side of the histograms.

\begin{figure}[h]
\begin{center}
    \includegraphics [scale=0.44] {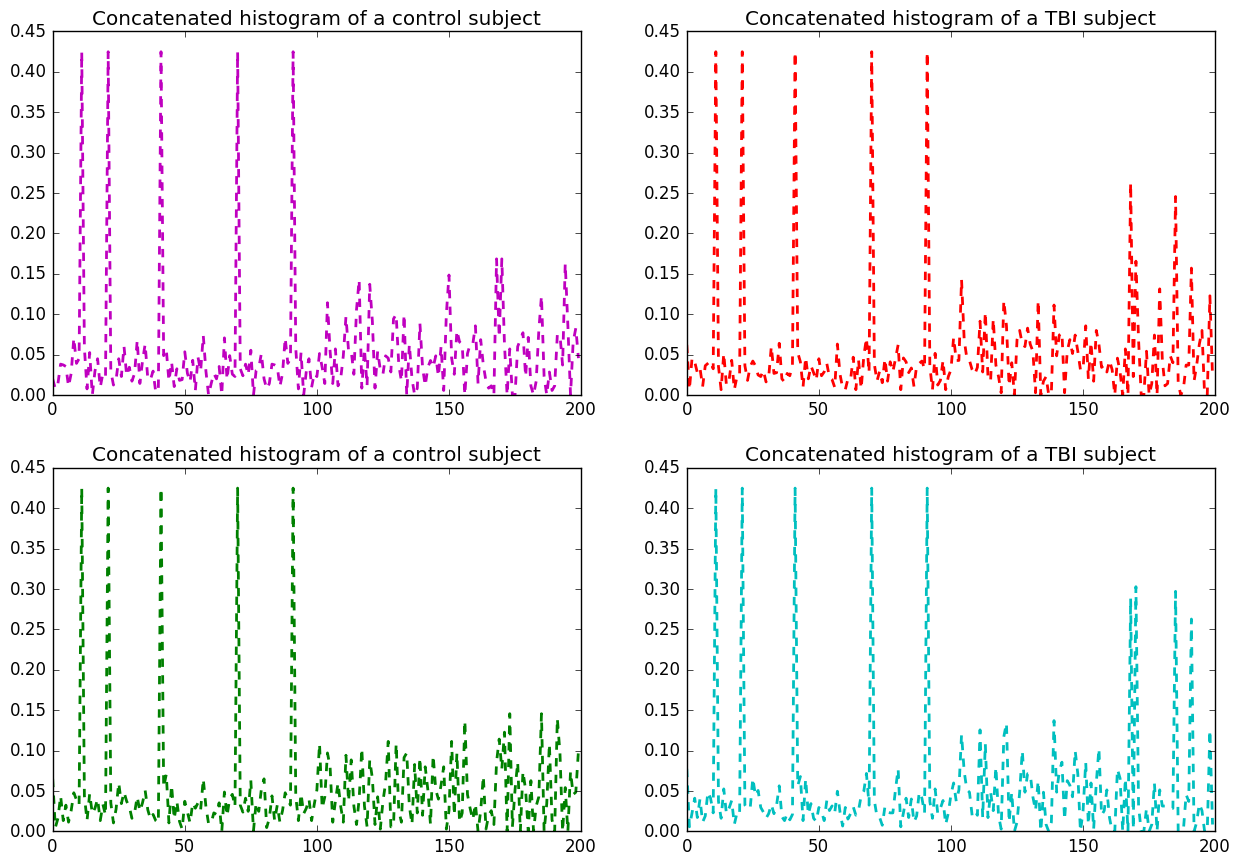}
\end{center}
\vspace{-0.2cm}
  \caption{Histogram comparison between two pairs of mTBI and control subjects.}
\end{figure}

\noindent
\\\textbf{Conclusions.}
Here we show the application of two approaches to multi-feature analysis using advanced diffusion MRI for classification of patients with mTBI compared with controls: first employing greedy forward feature selection and support vector machine algorithms and second exploring a novel approach in which computer vision and machine learning techniques are used to learn useful feature representation. Our early results show that feature learning approach achieves 5.5\% gain over mean value features.
These visual features can also be used for long-term outcome prediction of mTBI patients \cite{sherv}.

\end{document}